\documentclass[11pt]{article}
\usepackage{coling2016}
\usepackage{times}
\usepackage{url}
\usepackage{latexsym}
\usepackage{graphicx}
\usepackage{xcolor}
\usepackage[utf8]{inputenc}
\usepackage{dirtytalk}
\usepackage{enumitem}
\usepackage{caption}
\usepackage{subcaption}
\usepackage{epstopdf}
\usepackage{multirow}
\title{SentiHood: Targeted Aspect Based Sentiment Analysis Dataset for Urban Neighbourhoods}

\author{
  Marzieh Saeidi \\
  University College London \\
  {\tt msaeidi@cs.ucl.ac.uk} \\\And
  Guillaume Bouchard \\
  Bloomsbury AI \\
  {\tt guillaume@bloomsbury.ai} \\
  \AND
  Maria Liakata  \\
  University of Warwick \\
  {\tt m.liakata@warwick.ac.uk} \\\And
  Sebastian Riedel \\
  University College London \\
  {\tt sriedel@cs.ucl.ac.uk} \\
}

\begin{document}

\maketitle

\begin{abstract}
   In this paper, we introduce the task of targeted aspect-based sentiment analysis. The goal is to extract \emph{fine-grained} information with respect to entities mentioned in user comments. This work extends both \emph{aspect-based} sentiment analysis that assumes a single entity per document and \emph{targeted} sentiment analysis that assumes a single sentiment towards a target entity. In particular, we identify the sentiment towards each aspect of one or more entities. As a testbed for this task, we introduce the SentiHood dataset, extracted from a question answering (QA) platform where urban neighbourhoods are discussed by users. In this context units of text often mention several aspects of one or more neighbourhoods. This is the first time that a generic social media platform in this case a QA platform, is used for fine-grained opinion mining. Text coming from QA platforms is far less constrained compared to text from review specific platforms which current datasets are based on. We develop several strong baselines, relying on logistic regression and state-of-the-art recurrent neural networks.
\end{abstract}

\section{Introduction}
\label{intro}
\blfootnote{
    %
    %
    \hspace{-0.65cm}  
    This work is licensed under a Creative Commons Attribution 4.0 International Licence. Licence details: http://creativecommons.org/licenses/by/4.0/
    %
    %
    %
    %
}

    Sentiment analysis is an important task in natural language processing. It has received not only a lot of interest in academia but also in industry, in particular for identifying customer satisfaction on products and services. Early research in the field~\cite{das2001yahoo,morinaga2002mining} of sentiment analysis only focused on identifying the overall sentiment or polarity of a given text. The underlying assumption of this work was that there is one overall polarity in the whole text.
    
    \emph{Aspect-based} sentiment analysis (ABSA)~\cite{jo2011aspect,pontiki2015semeval,SemEval_2016_task5} relates to the task of extracting fine-grained information by identifying the polarity towards different aspects of an entity in the same unit of text, and recognizing the polarity associated with each aspect separately. The datasets for this task were mostly based on specialized review platforms such as Yelp where it is assumed that only one entity is discussed in one review snippet, but the opinion on multiple aspects can be expressed. This task is particularly useful because a user can assess the aggregated sentiment for each individual aspect of a given product or service and get a more fine-grained understanding of its quality.
    
    Another line of research in this field is \emph{targeted} (a.k.a. target-dependent) sentiment analysis~\cite{jiang2011target,vo2015target}. Targeted sentiment analysis investigates the classification of opinion polarities towards certain target entity mentions in given sentences (often a tweet). For instance in the sentence \say{People everywhere love Windows \& vista. Bill Gates}, polarity towards Bill Gates is \say{Neutral} but the positive sentiment towards Windows \& vista will interfere with identifying it if the usual methods for sentiment analysis task are employed. However this task assumes only the \emph{overall} sentiment for each entity. Moreover, the existing corpora for this task so far has contained only a single target entity per unit of text.
    
    Both settings are obviously limited, and there exists many scenarios in which sentiments towards different aspects of several entities are discussed in the same unit of text. As a running example, we use urban areas: choosing which area to live or to visit is an important task when moving or visiting a new city. Currently there are no dedicated platforms for reviewing and rating aspects of neighbourhoods of a city. However we can find many discussions and threads on several blogs and question answering platforms that discuss aspects of areas in many cities around the world. In general, these conversations are very comprehensible: they often contain specific information about several aspects of several neighbourhoods. One example is the following (area names are highlighted in bold and aspect related terms are underlined): \\
    
    \say{\textit{Other places to look at in South London are \textbf{Streatham} (good range of \underline{shops} and \underline{restaurants}, maybe a bit \underline{far out} of central London but you get more for your \underline{money}) \textbf{Brixton} (good \underline{transport links}, \underline{trendy}, can be a bit \underline{edgy}) \textbf{Clapham} (good \underline{transport}, good \underline{restaurants/pubs}, can feel a bit \underline{dull}, \underline{expensive}) ...}}\\
    
    The example above does not perfectly fit into the existing tasks in sentiment analysis mentioned earlier. In this work, we introduce a new task that not only subsumes the existing sub-fields of \emph{targeted} and \emph{aspect-based} sentiment analysis but it also makes less assumptions on the number of entities that can be discussed in the unit of text. 
    
    To compare with the existing aspect-based sentiment analysis task, take the following example from the restaurant dataset used by SemEval shared ABSA~\cite{SemEval_2016_task5} task. \say{\textit{The design of the \underline{space} is good but the \underline{service} is horrid!}}. The ABSA task aims to identify that a positive sentiment towards the \textit{ambiance} aspect is expressed (opinion target expression is \say{space}). Moreover, a negative sentiment is expressed towards the \textit{service} aspect (opinion target expression is \say{service}). In this example, it is assumed that both of these opinions are expressed about a single restaurant which is not mentioned explicitly. However, take the following \emph{synthetic} example that ABSA is not addressing: \\
    
    \say{\textit{The design of the \underline{space} is good in \textbf{Boqueria} but the \underline{service} is horrid, on the other hand, the staff in \textbf{Gremio} are very friendly and the \underline{food} is always delicious.}} \\
    
    In this example, more than one restaurant are discussed and restaurants for which opinions are expressed, are explicitly mentioned. We call these target entities. Current ABSA task can only recognise that positive and negative opinions towards aspect \say{service} are expressed. But it can not identify the target entity for each of these opinions (i.e. Germio and Boqueria respectively). Targeted aspect-based sentiment analysis handles extracting the target entities as well as different aspects and their relevant sentiments.
    
    In the following, we argue that this task is both very relevant in practice, and raises interesting modelling questions. To facilitate research on this task we introduce the SentiHood dataset. SentiHood is based on the text from a QA platform in the domain of neighbourhoods of a city. Table~\ref{tab:example_task} shows examples of input sentences and annotations provided.
    \begin{table}[ht]
        \centering
        \begin{tabular}{|l|l|}
        \hline
            \textbf{Sentence}                                                                    & \textbf{Labels}    \\
            \hline
            The cheap parts of London are \textbf{Edmonton} and \textbf{Tottenham} and they  & (Edmonton,price,Positive) \\
                    are all poor, crime ridden and crowded with immigrants      &  (Tottenham,price,Positive) \\
                        &  (Edmonton,safety,Negative) \\
                        & (Tottenham,safety,Negative)\\
            \hline
            \textbf{Hampstead} area, more expensive but a better quality of living than                    & (Hampstead,price,Negative)\\
             in \textbf{Tufnell Park}                                                          & (Hampstead,live,Positive)\\
            \hline
        \end{tabular}
        \caption{Examples of input sentences and output labels in the system.}
        \label{tab:example_task}
    \end{table}  
    
    Our contributions in this paper can be summarised as follows:
    \begin{itemize}
        \item We introduce the task of \emph{targeted aspect-based} sentiment analysis as a further step towards extracting more fine-grained information from more complex text in the field of sentiment analysis.
        \item We use the text from social media platforms, in particular QA, for fine-grained opinion mining. So far, all datasets in this field have utilised text from review specific platforms where certain assumptions can be made and data is more constrained and less noisy.
        \item We propose SentiHood, a benchmark dataset that is annotated for the task of targeted aspect-based sentiment analysis in the domain of urban neighbourhoods.
        \item We show that despite the fact that the texts in QA were not written with the goal of writing a review in mind, question answering platforms and online forums are in general rich in information.
        \item We provide strong baselines for the task using both logistic regression and Long Short Term Memory (LSTM) networks and analysis of the results.
    \end{itemize}
\section{SentiHood}
    SentiHood is a dataset for the task of \emph{targeted} \emph{aspect-based} sentiment analysis. It is based on the text taken from question answering platform of Yahoo! Answers that is filtered for questions relating to neighbourhoods of the city of London. In this section we explain the data collection and annotation process and summarise properties of the dataset.
    \subsection{Data Collection Process}
        Entities in the dataset are locations or neighbourhoods. Yahoo! Answers was queried using the name of each neighbourhood of the city of London. Location (entity) names were taken from the gazetteer GeoNames\footnote{\url{http://www.geonames.org/}} and restricted to those within the boundaries of London. This list includes names of areas and boroughs and therefore entities are not always geographically exclusive (a borough contains several areas or neighbourhoods). The content of each question-answer pairs was aggregated and split into sentences. We keep only sentences that have a mention of a location entity name and discard other sentences.
 
    \subsection{Categories}
    The Number of location mentions in a single sentence in our dataset varies from one to over $50$. To simplify the task, we only annotate sentences that contain one or two location mentions. These sentences were divided into two groups: sentences containing one location mention --- Single, and sentences containing two location mentions --- Multi. This is to observe the difficulty of annotating two groups by human annotators and by the models.
    
    \subsection{Aspects} 
    Like existing work in the aspect-based sentiment analysis task~\cite{brychcin2014uwb}, a pre-defined list of aspects is provided for annotators to choose from. These aspects are: \textit{live}, \textit{safety}, \textit{price}, \textit{quiet}, \textit{dining}, \textit{nightlife}, \textit{transit-location}, \textit{touristy}, \textit{shopping}, \textit{green-culture} and \textit{multicultural}. Adding an additional aspect of \textit{misc} was considered. However in the initial round of annotations, we realised that it had a negative effect on the decisiveness of annotators and it led to a lower overall agreement. Aspect \textit{general} refers to a generic opinion about a location, e.g. \say{I love \textbf{Camden Town}}.  
    
    \subsection{Sentiment}
    For each selected aspect, annotators were required to select a polarity or sentiment. Most work in this area considers three sentiment categories of \say{Positive}. \say{Negative} and \say{Neutral}. In our annotation however, we only provided \say{Positive} and \say{Negative} sentiment labels. This is because in our data we rarely come across cases where aspects are discussed without a polarity. 

    \subsection{Target Entity}
    Target entity is a location entity in which an opinion (aspect and sentiment) is expressed for. We also refer to target entity as target location.  
    
    \subsection{Out of scope}
    For the sentences that do not comply with our schema, we define the two following special labels. Sentences marked with one of the these labels are removed from the dataset.
    \begin{enumerate}
    \item \textbf{Irrelevant}: When the identified name does not refer to a location entity: for example in the sentence \say{\textbf{Notting Hill} (1999) stars Julia Roberts and Hugh Grant use the characteristic features of the area as a backdrop to the action}, \say{Notting Hill} refers to the movie and not the area.
    \item \textbf{Uncertain}: When two contradicting sentiments are expressed for the same location and aspect, e.g. \say{Like any other area, \textbf{Camden Town} has good and bad parts}. Moreover, when the opinion is expressed for an area without a direct mention in the sentences, e.g. \say{\textbf{It}'s a very trendy area and not too far from \textbf{King's Cross}}.
    \end{enumerate}

    \subsection{Procedure:} 
     We use the BRAT annotation tool~\cite{stenetorp2012brat} to simplify the annotation task. Three annotators were initially selected for the task. None of the annotators are experts in linguistics. Annotators began by reading the guidelines and examples. Each annotator was then required to annotate a small subset of the data. After each round of annotation, agreements between annotators were calculated and discussed and this procedure continued until they reached a reasonable agreement.
        10\% of the whole dataset was randomly selected and annotated by all the three annotators. The annotator with the highest inter-annotator agreement was selected to annotate all the dataset. 
        \paragraph{Agreements:} 
         Cohen's Kappa coefficient($K$)~\cite{cohen1960} is often used for measuring the pairwise agreement between each two annotators for the task of aspect-based sentiment analysis~\cite{gamon2005pulse,ganu2009beyond} and other tasks~\cite{liakata2010corpora}. The Kappa Coefficient is calculated over aspect-sentiment pairs per each location. Pairwise inter-annotator agreement for aspect categories measured using Cohen's Kappa is $0.73$, $0.78$ and $0.70$, which is deemed of sufficient quality. It is worth mentioning that agreements on different aspect categories varied, with some aspects having a higher agreement rate.
         Agreements for aspect expressions are $0.93$, $0.94$, $0.93$. These agreements indicate reasonably high inter-annotator agreements~\cite{pavlopoulos2014aspect}.
         
         \paragraph{Disagreements:}
         Main disagreements between annotators occurred in detecting the aspect rather than detecting the sentiment, aspect expression or the target location. For instance, some annotators associated the expression \say{residential area} with a \say{Positive} sentiment for aspect \say{quiet} or \say{live} and others did not agree that \say{residential} implies quietness or desirable for living. In the case of disagreements, the vote of the majority was considered as the correct annotation.
         
         Some ambiguity was also observed with respect to detecting the target location. This occurred mainly when a location is confined in another location. For instance the sentence \say{\textbf{Angel} in \textbf{Inslington} has many great restaurants for eating out} expresses a \say{Positive} sentiment for the aspect \say{dining} of area \textbf{Angel} which is within the borough of \textbf{Islington}. Some annotators suggested that the sentence also implies the same opinion for \textbf{Islington}. However at the end all annotators agreed that in such cases no implicit assumptions should be made and only confined area should be labeled.
         
        \subsection{Dataset}
\begin{figure}[ht]
          \centering
              \includegraphics[width=.8\linewidth]{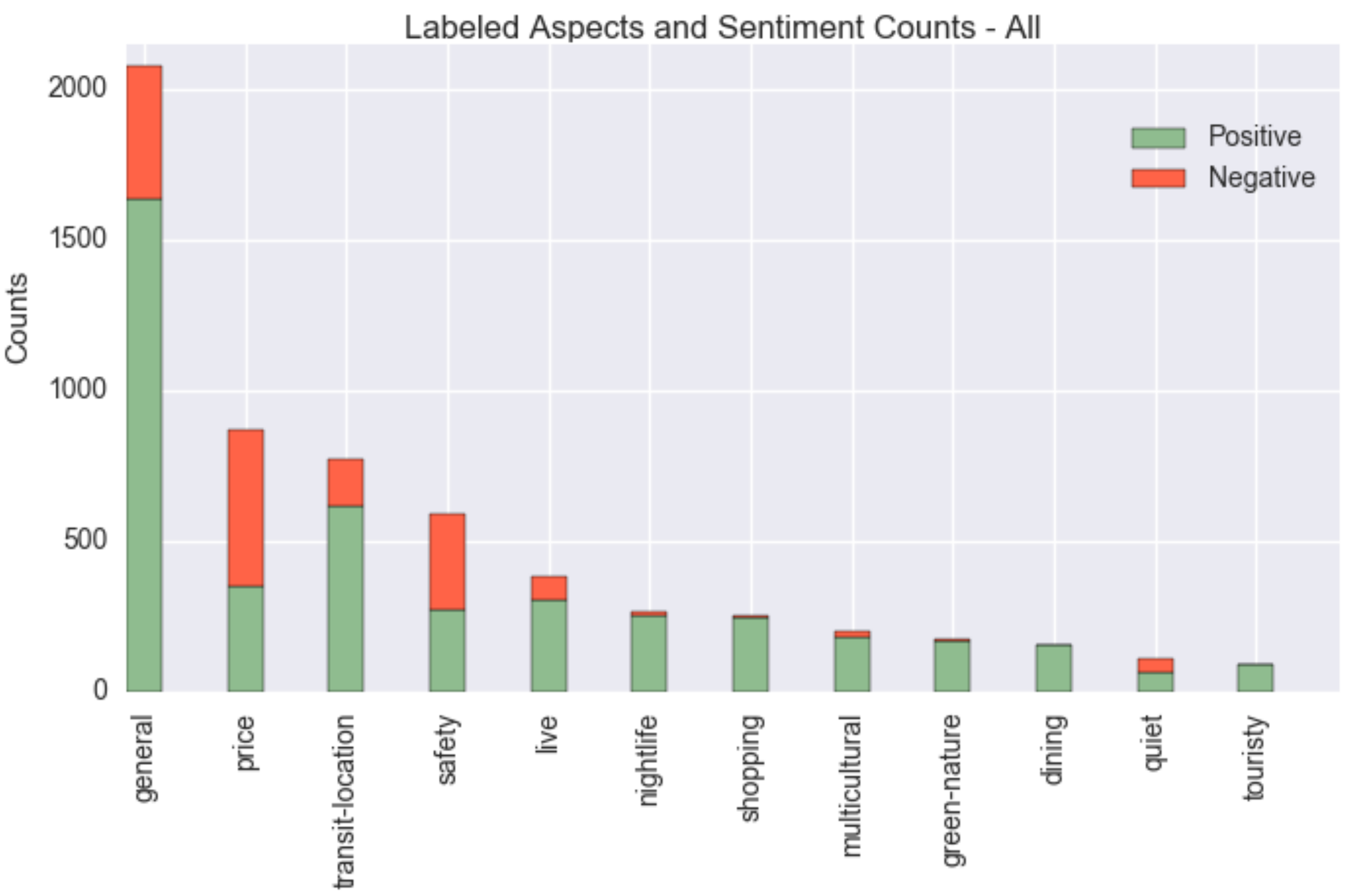}
          \caption{Number of annotated aspects and their sentiments.}
          \label{fig:sentihood_stat}
        \end{figure}

        SentiHood currently contains annotated sentences containing one or two location entity mentions.\footnote{SentiHood data can be obtained at \url{http://annotate-neighborhood.com/download/download.html}} SentiHood contains $5215$ sentences with $3862$ sentences containing a single location and $1353$ sentences containing multiple (two) locations. Figure~\ref{fig:sentihood_stat} shows the number of sentences that are labeled with each aspect, breaking down on the sentiment \say{Positive} or \say{Negative}. \say{Positive} sentiment is dominant for aspects such as dining and shopping. This shows that for some aspects, people usually talk about areas that are good for it as oppose to areas that are not. The \textit{general} aspect is the most frequent aspect with over $2000$ sentences while aspect \textit{touristy} has occurred in less than $100$ sentences. Notice that since each sentence can contain one or more opinions, the total number of opinions ($5920$) in the dataset is higher than the number of sentences.
        
        Location entity names are masked by \textbf{location1} and \textbf{location2} in the whole dataset, so the task does not involve identification and segmentation of the named entities. We also provide the dataset with the original location entity names. 
\section{Task}
        We define the task of targeted aspect-based sentiment analysis as follows: given a unit of text $s$ (for example, a sentence), provide a list of tuples (labels) $\{(l, a, p)\}_{t = 0}^{T}$, where $p$ is the polarity expressed for the aspect $a$ of entity $l$. Each sentence can have zero to $T$ number of labels associated with it.
        
        Within the current aspect-based sentiment analysis work, three tasks are defined~\cite{brychcin2014uwb}: detecting the aspect, detecting the opinion target expression and detecting the sentiment, with detecting the opinion target expression being an intermediary task for identifying the sentiment of the aspect.
        
        Here we focus on identifying only the aspect and sentiment for each entity. We identify each aspect, its relevant sentiment and the target location entity jointly by introducing a new polarity class called \say{None}. \say{None} indicated that a sentence does not contain an opinion for the aspect $a$ of location $l$. Therefore the overall task can be defined as a three-class classification task for each $(l,a)$ pair with labels \say{Positive}, \say{Negative}, \say{None}. Table~\ref{tab:example_task} shows an example of the input sentence and output labels. 
        \begin{table}[ht]
            \begin{center}
            \begin{tabular}{| l | c |}
            \hline
                \textbf{Sentence}                                                                    & \textbf{Labels}    \\
                \hline
                \multirow{2}{*}{\textbf{location1} is very safe and \textbf{location2} is too far}      & (location1,safety,Positive) \\
                                                                                                        & (location1,transit-location,None)\\
                                                                                                        & (location2,safety,None)\\
                                                                                                        & (location2,transit-location,Negative)\\
                \hline
            \end{tabular}
            \end{center}
            \caption{Example of an input sentence and the output labels.}
            \label{tab:example_task}
        \end{table} 
\section{Evaluation}
        Most existing work in aspect-based sentiment analysis field, report $F_1$ measure for aspect detection task, and accuracy for sentiment classification. The scores can be calculated over 2-class or 3-class sentiments~\cite{pontiki2015semeval}. In our results, $F_1$ score is calculated with a threshold that is optimized on validation set. 
        
        We also propose the AUC (area under the ROC curve) metric for both aspect and sentiment detection tasks. AUC captures the quality of the ranking of output scores and does not rely on a threshold. 
    
\section{Baseline}
        Here we propose baselines for the task. In all our methods, we treat the task as a three-class classification for each aspect and use a softmax function as follows:
        \begin{equation}\label{softmax}
        p(y_{l,a} = c) = \mathrm{softmax} (c) = \frac{exp(w_c . e_l + b_c)}{\sum_{c^\prime=1}^C exp(w_{c^\prime} . e_l + b_{c^\prime})}
        \end{equation}
        where $y_{l,a}$ is the sentiment label of aspect $a$ for location $l$. $w_c$ and $b_C$ are the weights and the bias specific to each sentiment class $c$, respectively. $e_l$ is a representation of location $l$. This representation can be a BoW or a distributional representation. Each method that we propose here define their own specific representation for $e_l$. 
    \subsection{Logistic Regression}
        Many existing works in the aspect-based sentiment analysis task,\footnote{including participants of SemEval ABSA tasks} use a classifier, such as logistic regression or SVM, based on linguistic features such as n-grams, POS information or more hand-engineered features. We can think of these features as a sparse representation $e_l$ that enter the softmax in equation \ref{softmax}. More concretely, we define the following sparse representations of locations:
        \paragraph{Mask target entity n-grams:} For each location, we define an n-gram representation over the sentence and mask the target location using a special token. This can help to differentiate between representations of two locations present in the same sentence.
        \paragraph{Left-right n-grams:} we create an n-gram representation for both the right and the left context around each location mention. We then concatenate these two representations to obtain one single feature vector.
        \paragraph{Left right pooling:} Previously embedding representations over the left and right context have been used for automatic feature detection in the targeted sentiment analysis task~\cite{vo2015target}.
        Inspired by this approach, we obtain max, min, average and standard deviation pooling over all the word embeddings for left and right context separately. We then combine the pooled embeddings of the left and right context to obtain a single feature vector. Word embeddings are obtained by running word2vec tool on a combination of our Yahoo! Answers corpus and a substantially big corpus from the web.\footnote{\url{http://ebiquity.umbc.edu/redirect/to/resource/id/351/UMBC-webbase-corpus}}
    \subsection{Long Short-Term Memory (LSTM)}
\begin{figure}[ht]
          \centering
              \includegraphics[width=.85\linewidth]{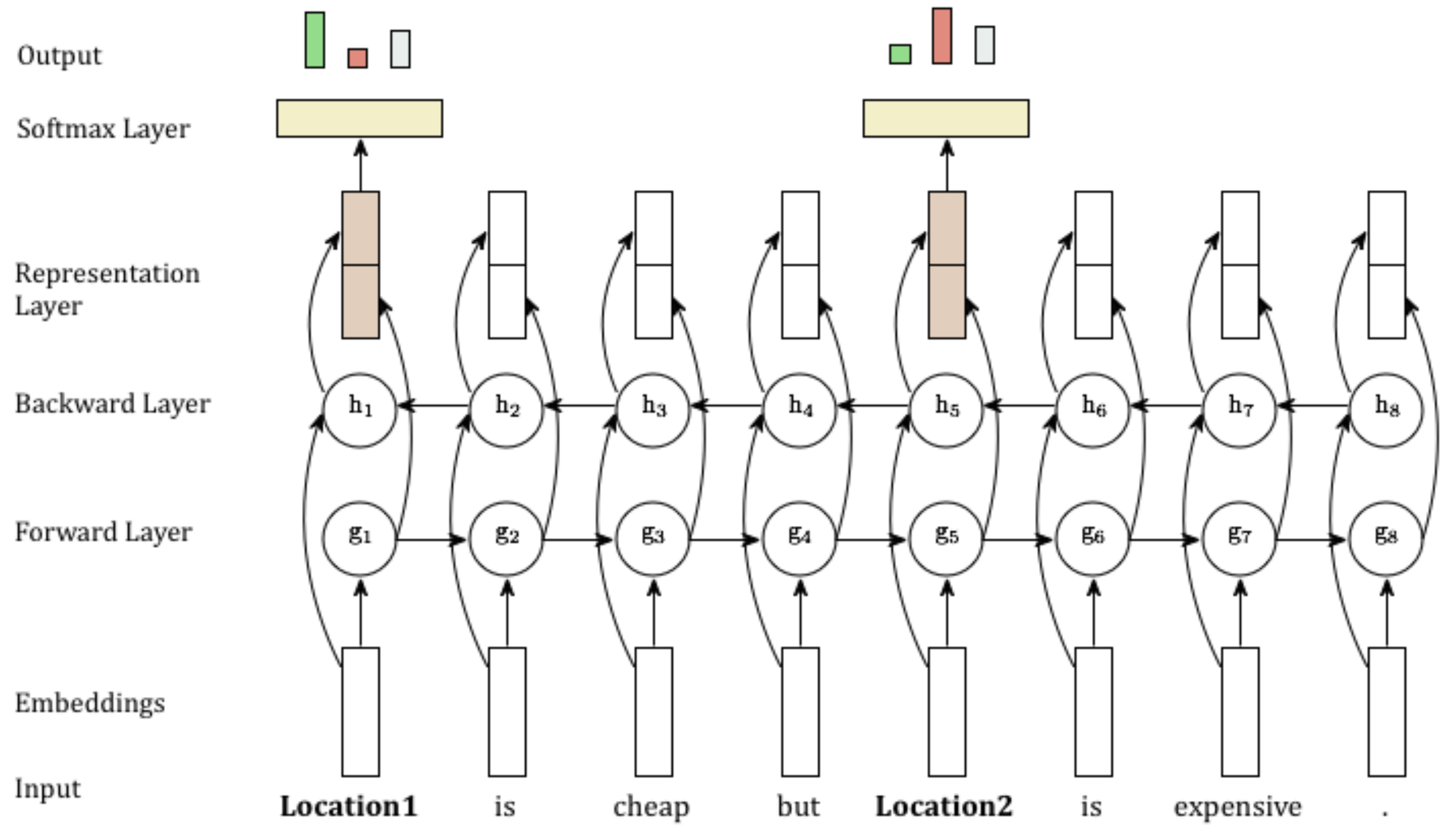}
          \caption{Bidirectional LSTM outputs a representation for each token in the sentence. The output state at the index of each location is then fed into a softmax layer to identify the sentiment class for the corresponding aspect. In this figure, LSTM is trained to identify the sentiment of aspect \say{price}. Model should predict \say{Positive} for \textbf{location1} and \say{Negative} for \textbf{location2}}
          \label{fig:model_bi_lstm}
        \end{figure}

        Inspired by the recent success of applying deep neural networks on language tasks, we use a bidirectional LSTM~\cite{hochreiter1997long} to learn a classifier for each of the aspects. Representations for a location ($e_l$) are obtained using one of the following two approaches:
        \paragraph{Final output state (LSTM - Final): } $e_l$ is the output embedding of the bidirectional LSTM.
        \paragraph{Location output state (LSTM - Location):} $e_l$ is the output representation at the index corresponding to the location entity as illustrated in Figure~\ref{fig:model_bi_lstm}.
        
\section{Experiments}
    In this paper, we select the four most frequent aspects from the dataset which are: \say{price}, \say{safety}, \say{transit-location} and \say{general} but the same approach can be applied to the remaining aspects. 
    We divide each collection of single and multiple location mentions into train, dev and test set, with each having $70\%$, $10\%$ and $20\%$ of data respectively. We choose the best model with respect to the dev set. 
    
    In the case of the LSTM, we evaluate the loss on both training set and dev set after each iteration. We save the best model which has the lowest loss on the dev set over all the iterations. We then run this model on the test set and report the results. 
    We report results separately on both categories of single location sentences and sentences with two locations and over all the data in the test set. Results on single location sentences mainly show the ability of the model to detect the correct sentiment for an aspect. On the other hand, results on two location sentences demonstrate the ability of the system not only on detecting the relevant sentiment of an aspect but also on recognising the target entity of the opinion.
    \paragraph{Training LSTMs} 
    We implement our LSTM models using tensorflow~\cite{tensorflow2015_whitepaper}. To tackle the problem of having an unbalanced dataset (i.e. too many \say{None} instances), we train the LSTM model in batches with every batch having the same number of sentences selected randomly from each sentiment class. We tune the hyper parameters of the model on the dev set. The best model uses hidden units of size $50$ and batch sizes of size $150$. The Adam optimizer is used for optimization with a starting learning rate of $0.01$ which is tuned to be the best performing on the dev set. Dropout is used both on initial word embeddings and on LSTM cells with the probability of $0.001$. Tensorflow~\cite{tensorflow2015_whitepaper} is used for the implementation of LSTM.
    \paragraph{Training Logistic Regression}
    Logistic regression models were based on implementations from scikit-learn.\footnote{\url{http://scikit-learn.org/}} Since we have an unbalanced dataset, we use a weighted logistic regression. To obtain the best weights, we cross-validate them on the development set. Weights inversely proportional to the size of each class result in the best performance.
\section{Results}
    Table~\ref{tab:results_models} shows the results (averaged over all selected aspects) in terms of both $F_1$/accuracy and AUCs. It also shows the results of logistic regression based models versus LSTM models. 
    
    As we can see, the n-gram representation with location masking achieves slightly better results over the left-right context. N-grams include unigrams and bigrams. Also, by adding POS information, we gain an increase in the performance. We also experimented with adding tri-grams but it did not have a positive effect on the overall scores. Separating the left and the right context (LR-Left-Right) for BoW representation, does not improve the performance. Left-right pooling of dense embeddings performed weakly in comparison with other representations and therefore their results were omitted. 

    Amongst the two variations of LSTM, the model with final state embeddings does slightly better than the model where we use the embeddings at the location index, however they are not significantly different (with a $p$ value less than $0.01$). It is interesting to note that the best LSTM model is not superior to logistic regression model, especially in terms of AUC. This can be due to the fact that the amount of training data is not sufficient for LSTM to perform well. Moreover, while we provide some grammar information to logistic regression model through POS tags, such information is not incorporated into LSTM models. Another interesting observation is that the $F_1$ measure for logistic regression model with n-grams and POS information is very low while this model's performance is superior to other models in terms of AUC. This is because in general, it is easier to rank prediction scores than to assign predicted labels to instances by choosing a hard threshold.
    \begin{table}[ht]
        \centering
        \begin{tabular}{l|r | r || r | r}
            Model    & Aspect ($F_1$) & Sentiment (Accuracy) & Aspect (AUC) & Sentiment (AUC) \\
            \hline
            LR-Left-Right&   $0.683$         &   $0.847$             &   $0.903$             &   $0.875$             \\
            LR-Mask(ngram)&   $\mathbf{0.697}$         &   $0.853$             &   $0.918$             &   $0.885$             \\
            LR-Mask(ngram+POS)&   $\mathit{0.393}$         &   $\mathbf{0.875}$    &   $\mathbf{0.924}$    &   $\mathbf{0.905}$    \\
            \hline      
            LSTM-Final&   $0.689$    &   $0.820$             &   $0.898$             &   $0.854$             \\
            LSTM-Location&   $\mathbf{0.693}$    &   $0.819$             &   $0.897$             &   $0.839$             \\
        \end{tabular}
        \caption{Results of best logistic regression (LR) models and LSTM models.}
        \label{tab:results_models}
    \end{table}
    
    Table~\ref{tab:results_categories} shows the average AUC (over aspect and sentimentclassification tasks) for two categories of data: Single --- sentences that contain one location entity and Multi --- sentences that contain two location entities. While logistic regression can perform slightly better on son Single location sentences, LSTM performs slightly better on Multi location sentences. 
    \begin{table}[ht]
        \centering
        \begin{tabular}{l| r | r}
            Model                       &           Single      &           Multi               \\
            \hline
            LR - Mask (n-gram + POS)           &   $\mathbf{0.916}$    &           $\mathbf{0.907}$    \\
            \hline      
            LSTM - Final                &       $0.872$         &           $0.890$             \\
        \end{tabular}
        \caption{Results of best logistic regression (LR) and LSTM models on sentences with a single location (Single) and multiple locations (Multi). AUC scores are averaged over aspect and sentiment classification tasks.}
        \label{tab:results_categories}
    \end{table}
    
    Table~\ref{tab:results_aspects} shows the break down of average AUC scores for each aspect. We can see that aspects such as \say{safety} can be predicted with a better AUC score than aspect \say{general}.  
    \begin{table}[ht]
        \centering
        \begin{tabular}{l| r | r | r | r }
            Model               &       Price           &   Safety              &   Transit             &           General         \\
            \hline
            LR - Mask (n-gram + POS)   &   $\mathbf{0.940}$    &   $\mathbf{0.960}$    &   $\mathbf{0.879}$    &           $0.864$         \\
            \hline
            LSTM - Final        &       $0.875$         &       $0.932$         &       $0.836$         &       $\mathbf{0.869}$     \\
        \end{tabular}
        \caption{LR and LSTM performance breakdown on aspects. AUC scores are averaged over aspect and sentiment detection.}
        \label{tab:results_aspects}
    \end{table}
    
    Table~\ref{tab:examples} shows examples of correct and incorrect predictions using the best logistic regression model. The top part of the table contains examples that each contain a single location entity. At the bottom of the table, a sentence with two location entities is provided. The system correctly identifies that a \say{Positive} sentiment is expressed for the \textit{general} aspect about \textbf{location2}. However, no sentiment is expressed for this aspect for \textbf{location1}.
    \begin{table}[h]
        \centering
        \begin{tabular}{| l | l | l | l |}
        \hline
             \textbf{Sentence}                                                              &    \textbf{Aspect} & \textbf{Predicted} & \textbf{Label}    \\
            \hline
            \textbf{location1} is not a nice cheap residential area to live trust me  	& Price & \textcolor{red}{Positive} & Negative \\
            i was born and raised there & & &\\
            \hline
            I think you'd find it tough to find something affordable 	&Price & \textcolor{red}{Positive} & Negative \\
             in \textbf{location1} & & & \\
             \hline
             I can't recommend \textbf{location1} for affordability & Price	 & \textcolor{blue}{Negative} & Negative \\
              & & & \\
             \hline
             \hline
             I only know about \textbf{location1}, most people prefer location2 & General & \textcolor{blue}{None} & None \\
              & & & \\
            \hline
             I only know about location1, most people prefer \textbf{location2} & General & \textcolor{blue}{Positive} & Positive \\
              & & & \\
            \hline
      \end{tabular}
        \caption{Examples of input sentences and predicted labels using the best system (LR - Mask (n-gram + POS). Target entity locations are highlighted in bold.}
        \label{tab:examples}
    \end{table} 
\section{Related Work}
    The term sentiment analysis was first used in~\cite{analyzer2003extracting}. Since then, the field has received much attention from both research and industry. Sentiment analysis has applications in almost in every domain and it raised many interesting research questions. Furthermore, the availability of a huge volume of opinionated data on social media platforms has accelerated the development in this area.
    
    In the beginning work on sentiment analysis mainly focused on identifying the overall sentiment of a unit of text. The unit of text varied from document~\cite{pang2002thumbs,turney2002thumbs}, paragraph or sentences~\cite{hu2004mining}. However, only considering the overall sentiment fails to capture the sentiments over the aspects on which an entity can be reviewed or sentiment expressed toward different entities. Two remedy this, two new tasks have been introduced: \emph{aspect-based} sentiment analysis and \emph{targeted} sentiment analysis.
    
    Aspect based sentiment analysis assumes a \emph{single entity} per a unit of analysis and tries to identify sentiments towards different aspects of the entity~\cite{lu2011multi,lakkarajuaspect,alghunaim2015vector,bagheri2013care,somprasertsri2008automatic,alghunaim2015vector,lu2011multi,titov2008modeling,brody2010unsupervised}. This task however does not consider more than one entity in the given text. 

    Targeted (target dependent) sentiment analysis is another task that identifies polarity towards a target entity (as opposed to over entire unit of text)~\cite{mitchell2013open,jiang2011target,dong2014adaptive,vo2015target,zhang2016gated}.~\cite{jiang2011target} was the first to propose targeted sentiment analysis on Twitter and demonstrates the importance of targets by showing that 40\% of sentiment errors are due to not considering them in classification. However this task only identifies the \emph{overall sentiment} and the existing corpora for the task consist only of text with one single entity per unit of analysis.

    The task of targeted aspect-based sentiment analysis caters for more generic text by making fewer assumptions while extracting fine-grained information.
    
\section{Conclusion}
In this paper, we introduced the task of \emph{targeted} \emph{aspect-based} sentiment analysis and a new dataset. We also provide two strong baselines using logistic regression and LSTM. Ways to improve the baselines can involve using parse trees for identifying the context of each location. Data augmentation can be used to make the models and especially LSTM more robust to variations in the data. We also like to provide more detailed analysis of what each system can achieve.

\bibliographystyle{acl}
\bibliography{sentihood}

\end{document}